\documentclass[sn-mathphys,Numbered]{sn-jnl}

\usepackage{graphicx}
\usepackage{dsfont}
\usepackage{amsmath}
\usepackage{float}
\usepackage{mathtools}
\usepackage[T1]{fontenc}
\usepackage{amsmath}
\usepackage{caption}
\usepackage{subcaption}
\usepackage{textalpha}
\usepackage{dsfont}
\usepackage{amsmath}
\usepackage{float}
\usepackage{multirow}
\usepackage{hhline}
\usepackage{longtable}
\usepackage{caption}
\usepackage{xcolor,pifont}
\usepackage{multicol}
\usepackage{graphicx}%
\usepackage{multirow}%
\usepackage{amsmath,amssymb,amsfonts}%
\usepackage{amsthm}%
\usepackage{mathrsfs}%
\usepackage[title]{appendix}%
\usepackage{xcolor}%
\usepackage{caption}
\usepackage{textcomp}%
\usepackage{manyfoot}%
\usepackage{booktabs}%
\usepackage{algorithm}%
\usepackage{algorithmicx}%
\usepackage{algpseudocode}%
\usepackage{listings}%
\usepackage{adjustbox}
\usepackage{tabularx}
\usepackage{booktabs}
\usepackage{colortbl}
\usepackage{adjustbox}
\usepackage{microtype}
\usepackage{multirow}
\usepackage{pifont}
\usepackage{xcolor}
\newcommand{\xmark}{\ding{55}}%
\newcommand*\colourcheck[1]{%
  \expandafter\newcommand\csname #1check\endcsname{\textcolor{#1}{\ding{52}}}}
\colourcheck{blue}
\colourcheck{green}
\colourcheck{red}

\begin{document}

\title[Article Title]{End-to-End Semi-Supervised approach with Modulated Object Queries for Table Detection in Documents}


\author[1]{ \sur{Iqraa Ehsan}}
\author*[1,2,3]{\sur{Tahira Shehzadi}} \email{tahira.shehzadi@dfki.de}
\author[1,2,3]{\sur{ Didier Stricker}}
\author[1,2,3]{\sur{ Muhammad Zeshan Afzal}}

\affil[1]{\orgdiv{Department of Computer Science}, \orgname{Technical University of Kaiserslautern}, \orgaddress{ \country{Germany}}}
\affil[2]{\orgdiv{Mindgarage}, \orgname{Technical University of Kaiserslautern}, \orgaddress{\country{Germany}}}
\affil[3]{\orgname{German Research Institute for Artificial Intelligence (DFKI)}, \orgaddress{ \city{Kaiserslautern}, \postcode{67663}, \country{Germany}}}


\abstract{
Table detection, a pivotal task in document analysis, aims to precisely recognize and locate tables within document images. Although deep learning has shown remarkable progress in this realm, it typically requires an extensive dataset of labeled data for proficient training. Current CNN-based semi-supervised table detection approaches use the anchor generation process and Non-Maximum Suppression (NMS) in their detection process, limiting training efficiency. Meanwhile, transformer-based semi-supervised techniques adopted a one-to-one match strategy that provides noisy pseudo-labels, limiting overall efficiency. This study presents an innovative transformer-based semi-supervised table detector. It improves the quality of pseudo-labels through a novel matching strategy combining one-to-one and one-to-many assignment techniques. This approach significantly enhances training efficiency during the early stages, ensuring superior pseudo-labels for further training. Our semi-supervised approach is comprehensively evaluated on benchmark datasets, including PubLayNet, ICADR-19, and TableBank. It achieves new state-of-the-art results, with a mAP of 95.7\% and 97.9\% on TableBank (word) and PubLaynet with 30\% label data, marking a 7.4 and 7.6 point improvement over previous semi-supervised table detection approach, respectively. The results clearly show the superiority of our semi-supervised approach, surpassing all existing state-of-the-art methods by substantial margins. This research represents a significant advancement in semi-supervised table detection methods, offering a more efficient and accurate solution for practical document analysis tasks.}

\keywords{Table Detection, Document Analysis, Semi-Supervised Learning, Detection Transformer}



\maketitle
\vspace{-2em} 
\section{Introduction}\label{sec1}
A visual summary is crucial for many applications, such as document summaries and identifying graphical components in the visualization pipeline. Therefore, a crucial stage in the summary and analysis of the document will be localizing and identifying graphical elements. Tables are one of the integral elements in document analysis, adeptly consolidating crucial information into a format that maximizes spatial efficiency. Precise text extraction, especially within intricate table layouts, remains a recurring challenge. Manually extracting table objects has become impractical due to the growing number of documents. Automated methodologies \cite{mori1992historical, nguyen2021survey} offer dependable, impactful, and streamlined resolutions for manual tasks. OCR \cite{singh2012survey} is of paramount importance in the field of document analysis, particularly in the context of tables. It plays a crucial role in addressing the complexities associated with handling structured data. Conventional OCR\cite{bourbakis1991ocr} systems often falter when confronted with the structured format of tables, leading to misinterpretations, data loss, and formatting disparities. Early attempts \cite{mori1992historical} to address these challenges relied on external metadata embedded within documents. However, these methods proved inadequate for adapting to novel table configurations, such as borderless tables. 

Recent advancements in deep learning methodologies have transformed table detection \cite{bhatt2018optical, prasad2020cascadetabnet, shekar2021optical,shehzadi2023object5}, notably through the integration of object detection concepts from computer vision. As technology advances, new challenges have emerged within the domain of table detection. A notable challenge is that training deep learning models necessitates extensive annotation of a substantial amount of data, a task that demands substantial resources. While supervised approaches \cite{xue2022language, amrhein2018supervised} excel on standardized benchmarks, their application in industrial contexts is according to the availability of domain-specific annotated datasets. The process of manually generating or preprocessing labeled data is not only time-consuming but also introduces a risk of inaccuracies. As a result, there's a growing demand for networks that can effectively function with limited labeled datasets. 

Semi-supervised learning presents an invaluable solution precisely in situations where labeled data is limited. These techniques utilize both labeled and unlabeled data, subsequently providing pseudo-labels for the unlabeled instances. The goal of semi-supervised techniques is to improve model performance and generalization abilities by utilizing the advantages of both types of data. By doing so, it harnesses the inherent patterns and structures in the unlabeled data, enhancing the model's learning capacity and potentially improving its accuracy, all while reducing the dependency on extensive labeled datasets. Earlier, Semi-Supervised Object Detection (SSOD) \cite{softTeacher56} relies heavily on object detectors containing anchor generation process and non-maximum suppression (NMS) \cite{hosang2017learning} for post-processing. Later on, DETR \cite{detr34} has been adapted for semi-supervised settings, capitalizing on its inherent design which eliminates the need for NMS and efficiently utilizes both labeled and unlabeled data. The integration of semi-supervised learning in DETR presents challenges. DETR's dependence on a fixed set of object queries, established during training, may lead to inefficiencies in scenes with varying object numbers, potentially causing missed detections or false positives. Furthermore, the one-to-one assignment can exacerbate errors when pseudo-labels are inaccurate. For instance, if a pseudo-label, though not entirely precise, is incorporated into the training process, the one-to-one assignment scheme may compel a predicted bounding box to closely align with this inaccurate pseudo-label.

In response to the challenges addressed, we introduce a novel semi-supervised framework based on the Teacher-Student model. Our innovative matching module integrates two training stages, utilizing one-to-many \cite{fang2023feataug} and one-to-one \cite{li2023one} assignment strategies. The one-to-one matching ensures NMS-free end-to-end detection  \cite{hosang2017learning} advantages. In contrast, one-to-many matching generates higher-quality pseudo-labels for unlabeled data, contributing to a more efficiently optimized detector. In our semi-supervised learning approach, the student and teacher modules are equipped to handle both one-to-many \cite{fang2023feataug} and one-to-one \cite{li2023one} queries. This unified processing fed object queries from the weakly augmented images to the teacher module and those from strongly augmented images to the student module. Furthermore, we implement augmented ground truth in the one-to-many matching scheme, generating more high-quality pseudo-labels.

In summary, this paper presents significant contributions in the following:
\begin{itemize}
   \item We introduce a novel end-to-end DETR-based semi-supervised approach for table detection. The approach utilizes a dual query assignment mechanism that integrates the one-to-many matching strategy with the one-to-one matching strategy in the training phase.

    \item We introduce a query filtering module in a one-to-many matching strategy to provide high-quality pseudo-labels. To the best of our knowledge, this research is the first to focus on object queries in a semi-supervised setting for table detection tasks.
 
  \item We conduct a comprehensive evaluation using diverse datasets, including PubLayNet \cite{PubLayNet3}, ICDAR-19 \cite{icdar19}, and TableBank \cite{tablebank8}. Our approach surpasses both CNN-based and transformer-based semi-supervised approaches due to its superior efficiency and ability to generate high-quality pseudo-labels without using Non-maximum Suppression (NMS), resulting in a comparable performance with improved accuracy.
\end{itemize}

The subsequent sections of the paper are organized in the following manner. Section ~\ref{sec:related_work} provides an overview of prior research conducted in the field of detecting table objects inside document images.
Section~\ref{sec:method} explains the approach, while Section~\ref{sec:Experimental} presents an overview of the dataset utilized, outlines the evaluation criteria, and provides implementation details. We examine the experimental outcomes in Section ~\ref{sec:Results}. The ablation study is covered in Section~\ref{sec:Ablation}. Finally, Section ~\ref{sec:Conclusion} provides a comprehensive overview of the experimental work conducted and offers insights into potential future research topics.  
\vspace{-0.5em} 
\begin{figure}
    \centering
    \includegraphics[width=0.7\linewidth]{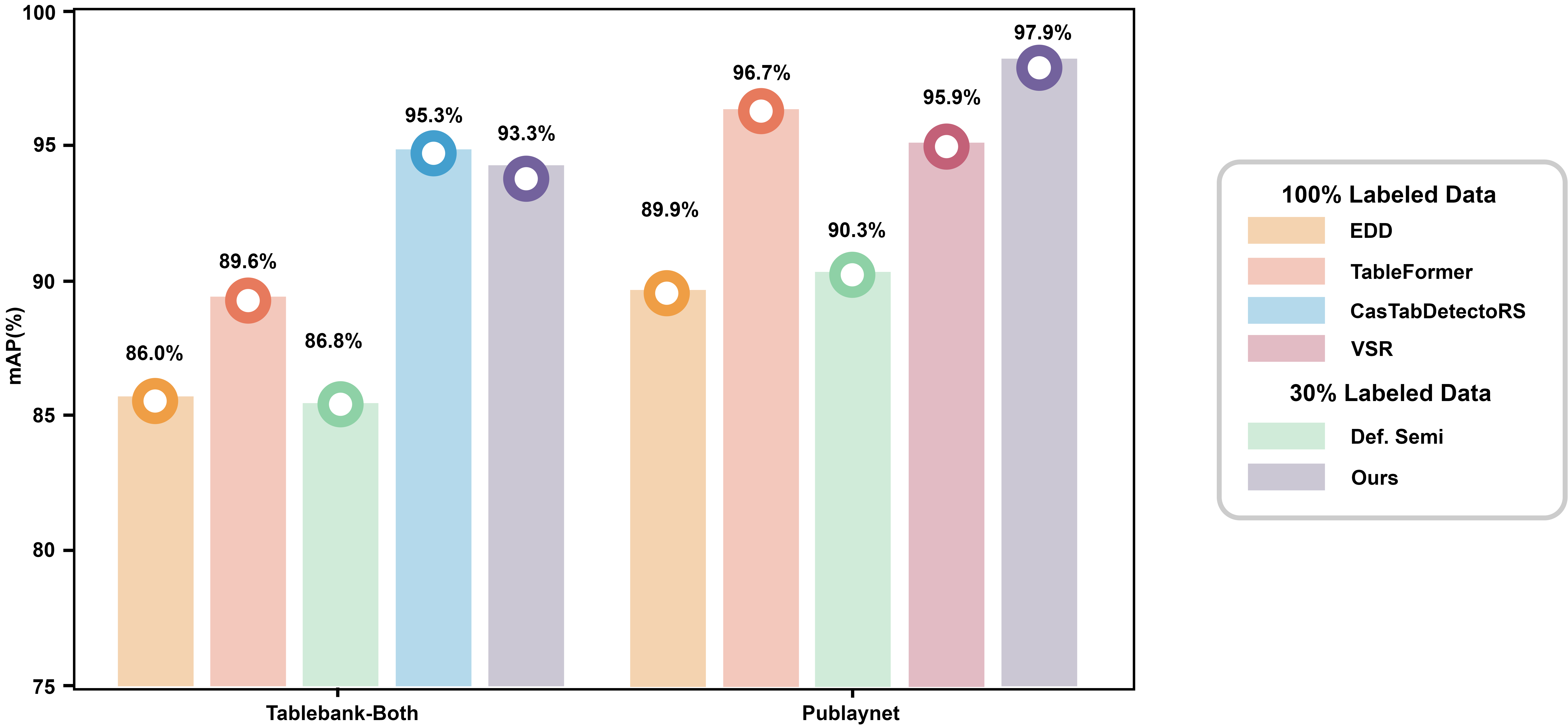}
    \captionsetup{font=small}
    \caption{Performance comparison of our approach with previous supervised and semi-supervised table detection approaches on TableBank-Both and Publaynet datasets. In a semi-supervised setting, we perform experiments with a 30$\%$ label data. The semi-supervised deformable transformer is referred to as the Def. Semi \cite{shehzadi_semi-detr_table1}. For an extensive summary of the results, please refer to  Table~\ref{tab:tableno8}.}
     \label{fig: Figure: 1}
     \vspace{-1em} 
\end{figure}

\section{Related Work}
\label{sec:related_work}
The identification and extraction of tables in document image analysis is a crucial and significant task. Numerous studies have introduced various methods for identifying tables with diverse structures in document images. In the past, a significant number of these methodologies either depended on pre-established rules or required further meta-data input to tackle table detection tasks \cite{tupextract3,DPmatch4}. In recent times, researchers have increasingly utilized statistical methodologies \cite{kieninger1998table} and included deep learning techniques to augment the flexibility and adaptability of table detection systems \cite{DeepDeSRT3,CasTab45,Hyb65}. This section provides an in-depth overview of these methodologies and offers insights into the strategies used for semi-supervised object detection based on Convolutional Neural Networks (CNNs).

\subsection{Rule-based Strategies}
Itonori et al. \cite{tsk5} pioneered table detection in document images by representing tables as text blocks with predefined rules. Subsequent studies by the same authors introduced a method considering both horizontal and vertical lines \cite{strctRTB3}. Pyreddy et al. \cite{TINTIN67} used tailored heuristics for tabular section detection, while Pivk et al. \cite{PIV67} transformed HTML formatted tables into logical forms. Hu et al. \cite{hu1999medium} proposed a method based on identifying white regions and vertically connected elements. Rule-based methods, effective for specific formats, are discussed in \cite{RecogTable5,extractTab9,TSsurvey32}. While effective, rule-based strategies may lack universal applicability, highlighting the need for more adaptable systems in the challenge of detecting tables in document images.

\subsection{Learning-Based Methodologies}
Cesarini et al.'s \cite{trainTD5} supervised learning approach is intended to reliably detect and categorize table elements within document images. It entails converting an image of a document into an MXY tree structure, then recognizing and categorizing individual components as tables according to the presence of vertical and horizontal lines. To detect tables, researchers utilize Hidden Markov Models \cite{RichMM3,FTabA4} and a Support Vector Machine (SVM) classifier with traditional heuristics~\cite{lineln6}. Recent advancements in deep learning methodologies have broadened their applications, extending from healthcare~\cite{shehzadi_IEEE_I9,Protein10}, traffic analysis~\cite{wajahatCC8}, to document analysis~\cite{continuaLR45,Real_DICls4,rethink78,naik86,cas10}. While machine learning outperforms rule-based approaches, it depends on additional information like ruling lines. In contrast, deep learning-based methodologies show higher precision and effectiveness. These techniques can be categorized into object identification, semantic segmentation, and bottom-up approaches.

\noindent\textbf{Approaches based on semantic segmentation.}
Table detection is framed as a segmentation problem in techniques like \cite{Xi17,He761,Ik36,Paliw9}, leveraging existing semantic segmentation networks to create pixel-level segmentation masks. These masks are then consolidated for the ultimate table detection results, showcasing superior performance over conventional methods on benchmark datasets \cite{icdar19,PubLayNet3,tablebank8,pubtables5}. In Yang et al.'s work \cite{Xi17}, a fully convolutional network (FCN) \cite{FCNseg4} integrates linguistic and visual features, enhancing segmentation for tables and other page elements. In order to generate segmentation masks for text and table sections as well as their outlines, He et al. \cite{He761} suggested a multi-scale FCN that was further improved for the final table elements.

\noindent\textbf{Inductive Approach}
The approaches employed in this study consider identifying tables as a problem of graph labeling, whereby the nodes of the graph represent various page components and the edges indicate the relationships between them. In their study, Li et al. \cite{Li64} employed a conventional layout analysis approach to extract line regions from a given document. These line areas were further categorized into four distinct classes, including text, figure, formula, and table. This classification process was accomplished through the utilization of two CNN-CRF networks. The researchers made further predictions on the cluster that would correlate to pairs of line areas. In contrast, Holecek et al. \cite{Martin66} addressed the analysis of text sections by seeing them as individual nodes within a graph structure, enabling the identification and understanding of the overall document architecture. Graph-neural networks were later utilized for node-edge categorization. It is important to acknowledge that these methodologies are contingent upon certain assumptions, such as the incorporation of text line boxes as supplementary input. 

\noindent\textbf{Approaches Based on Object Detection.}
The identification and localization of tables inside document images have often been addressed as a task of object detection, initially employing R-CNN methods like those by Yi et al. \cite{Yi77} and Hao et al. \cite{Hao789}, relying on heuristic rules. Advancements introduced efficient single-stage \cite{retinaNet68}, \cite{yolos6} and two-stage \cite{fast15}, \cite{faster23}, \cite{mask86}, \cite{cascadercnn8} detectors for document elements~\cite{shehzadi2024hybrid6,shehzadi2024endtoend7,EmmDocClassifier7}. Some studies \cite{Azka62,Ayan29,arif48} applied image transformations for improvement, while Siddiqui et al. \cite{Sidd32} enhanced Faster R-CNN with deformable-convolution and RoI-Pooling for geometrical modifications. The researchers Agarwal et al. \cite{Agarwal52} employed a composite network using deformable convolution to augment the performance of Cascade R-CNN. In contrast, the presented semi-supervised technique, treating detection as a set prediction task \cite{bridging_per3}, eliminates stages like proposal generation and post-processing (e.g., NMS \cite{hosang2017learning}), simplifying and streamlining the detection process for increased efficiency.

\subsection{Object Detection with Semi-supervised}
In semi-supervised object detection, two main methods emerge: consistency-based \cite{jeong2019consistency,propsemi8} and pseudo-label generation-based \cite{omnisup8,zoph2020rethinking,semimask4,sparse_semi_detr2,simplesemi76}. Prior research \cite{omnisup8,zoph2020rethinking} has used diverse techniques to generate pseudo-labels, like combining prediction results or using models like SelectiveNet \cite{selfsup6} to transfer bounding boxes from unannotated to annotated data. Research shows the superiority of self-training over pre-training, as demonstrated by STAC \cite{wei2021stac} using hard pseudo-labels. However, relying solely on initial pseudo-label predictions may hinder model accuracy improvement. Approaches like Unbiased Teacher \cite{unbiasedT36} use a two-stage framework to enhance quality via techniques like exponential moving average (EMA) and focal loss. Methods such as Mean Teacher \cite{tarvainen2017mean}, Soft Teacher \cite{softTeacher56}, and Instant-Teaching \cite{zhou2021instant} employ CNN-based teacher-student models for semi-supervised object detection. These methods vary in their approaches, from weighting classification losses based on teacher network outputs to employing instant pseudo-labeling schemes with extended data augmentations. However, these techniques integrate NMS into their detection process, potentially limiting training efficiency. Omni-DETR \cite{wang2022omni} and Semi-Supervised Table Detection \cite{shehzadi_semi-detr_table1} focus on stable and precise pseudo-label generation, improving the generalization capability of detectors. However, the use of one-to-one matching may introduce limitations that affect its ability to accurately recognize various objects, potentially impacting its overall performance. Our method enhances semi-supervised learning by eliminating traditional post-processing steps like NMS while improving pseudo-label quality, leading to improved model performance with limited data.

\section{Method}
\label{sec:method}
\subsection{Preliminary}
The DETR model \cite{detr34}, which is commonly used for applications such as object detection, consists of transformer encoder-decoder network that provides object class and bounding box position\cite{detr34, vaswani2017attention} as output. The process initiates by passing an input image through the backbone and transformer encoder \cite{vaswani2017attention}, resulting in the extraction of image features. The decoder subsequently takes the object queries \(Q\) which is written as \(\left\{ q_{1}, q_{2}, ..., q_{N} \right\}\) and image features. The DETR model utilizes task-specific prediction heads on the embeddings of object queries (\(\tilde{Q}\)) at every transformer decoder layer. These prediction heads are then followed by the predictors, resulting in an output implied as \(Y = \left\{ y_{1}, y_{2}, ..., y_{N} \right\}\).

The decoder employs a self-attention mechanism on object queries, facilitating interactions between queries to gather information relevant to duplicate detection. Then, it employs cross-attention between queries and image features to extract relevant data from the image features for object detection. Lastly, it incorporates a feed-forward network that independently analyzes the queries for final detection.

DETR ~\cite{detr34} generates a fixed number of predictions for each image, regardless of the actual number of objects present. The DETR employs a one-to-one strategy, establishing a bipartite matching between the predictions and the ground-truth objects:
\vspace{-0.5em} 
\begin{equation}
\left( y_{\sigma\left( 1 \right)}, \bar{y}_{1} \right), \left( y_{\sigma\left( 2 \right)}, \bar{y}_{2} \right), ..., \left( y_{\sigma\left( N \right)}, \bar{y}_{N} \right)
\end{equation}

In this case, the optimal permutation of \(N\) indices is denoted by \(\sigma\left( \cdot  \right)\), whereas the ground truth is denoted by  \(\bar{Y} = \left\{ \bar{y}_{1}, \bar{y}_{2}, ..., \bar{y}_{N} \right\}\). Hence, the expression of the loss function is formulated as:
\vspace{-0.5em} 
\begin{equation}
\mathcal{L} = \sum_{n=1}^{N} \mathscr{l}\left( y_{\sigma\left( n \right)}, \bar{y}_{n} \right)
\end{equation}
where $\mathcal{L}$ represents the loss between the predicted value \(y\) and the ground truth object \(\bar{y}\). The one-to-one assignment strategy in DETR matches each prediction to a unique ground truth, reducing duplicates and simplifying object detection. This strategy simplifies the detection process and improves accuracy by eliminating the need for post-processing steps like Non-Maximum Suppression (NMS), commonly used in object detection models to handle overlapping detections. However, the one-to-one matching in DETR limits learning to a small number of positive samples per image, potentially affecting its ability to recognize various objects and impacting overall performance.

\subsection{Dual Assignment Strategies}
The one-to-one query assignment strategy for table detection faces challenges such as mismatches between queries and tables, leading to inefficiencies and inaccuracies in detection. It can result in unassigned queries or undetected tables, particularly in documents with a variable number of tables. This issue becomes more prominent in semi-supervised settings, where label data is limited. It needs to generate pseudo-labels for unlabeled data. However, matching these noisy pseudo-labels with object queries employing a one-to-one matching strategy results in low performance due to inaccurate pseudo-labels~\cite{shehzadi_semi-detr_table1}.

\begin{figure}
    \centering
    \includegraphics[width=\linewidth]{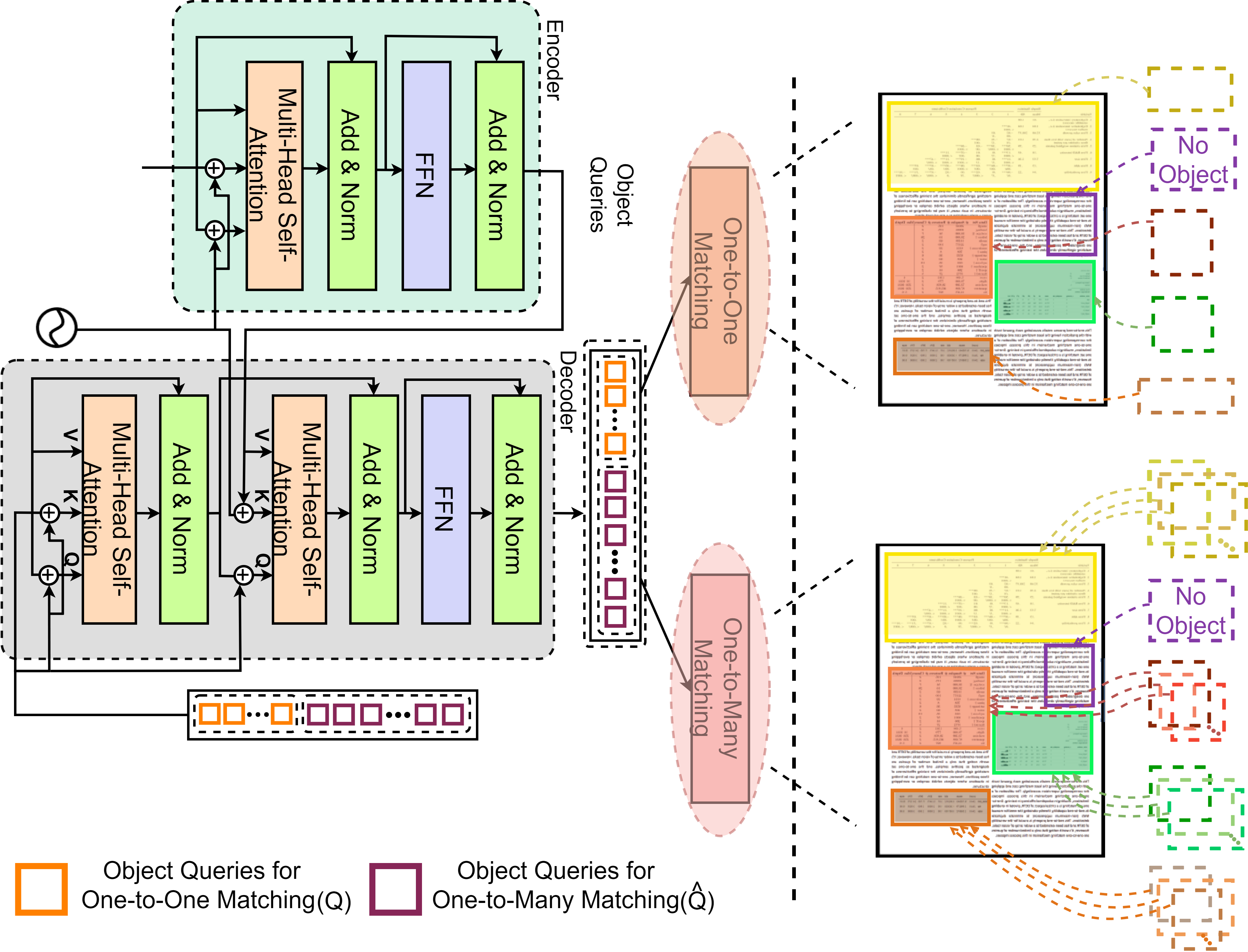}
    \captionsetup{font=small}
    \caption{ Our approach contains two modules, the student and teacher module. The decoder of the student-teacher modules utilize both one-to-one and one-to-many assignment strategies. The decoder of the teacher module employs the one-to-many assignment strategy to generate high-quality pseudo-labels and improve performance with limited data. The one-to-many assignment strategy in the student module provides high-quality predictions, while the one-to-one matching strategy removes duplications.}
     \label{fig: Figure: 2}
     \vspace{-1em} 
\end{figure}

To address the limitations of one-to-one matching in semi-supervised setting, our approach combines both one-to-one and one-to-many matching strategies as shown in Fig.~\ref{fig: Figure: 2}. This approach combines the strengths of both strategies to improve the overall efficiency of the object detection model. Within the matching strategies, two query sets are employed: $Q=\left\{q_{1}, q_{2}, · · ·, q_{N} \right\}$ and $\hat{Q}=\left\{ \hat{q_{1}}, \hat{q_{2}}, · · ·, \hat{q_{T}} \right\}$. Subsequently, predictions based on $Q$ or $\hat{Q}$ undergo either one-to-one or one-to-many matching strategies.

In one-to-one matching, the initial query set, \(Q\), is fed to the transformer decoder layers \cite{vaswani2017attention}, generating predictions for each layer's output. For this, bipartite matching is conducted between the pairs of predictions and ground truth across every layer. The loss is then determined using the formula:
\vspace{-1em} 
\begin{equation}
\vspace{-0.8em} 
\mathcal{L}_{o2o} = \sum_{l=1}^{L} \mathcal{L}_{H}\left( {y}_{p}, y_{g}\right)
\end{equation}
Here, \({y}_{p}\) signifies the outcomes generated by the transformer decoder layer at position $l$, while \(y_{g}\) denotes the ground-truth. The optimal assignment is efficiently calculated through the application of the Hungarian algorithm. One-to-one matching strategies are applied during both the model's training and testing phases.

\begin{figure}
    \centering
    \includegraphics[width=\linewidth]{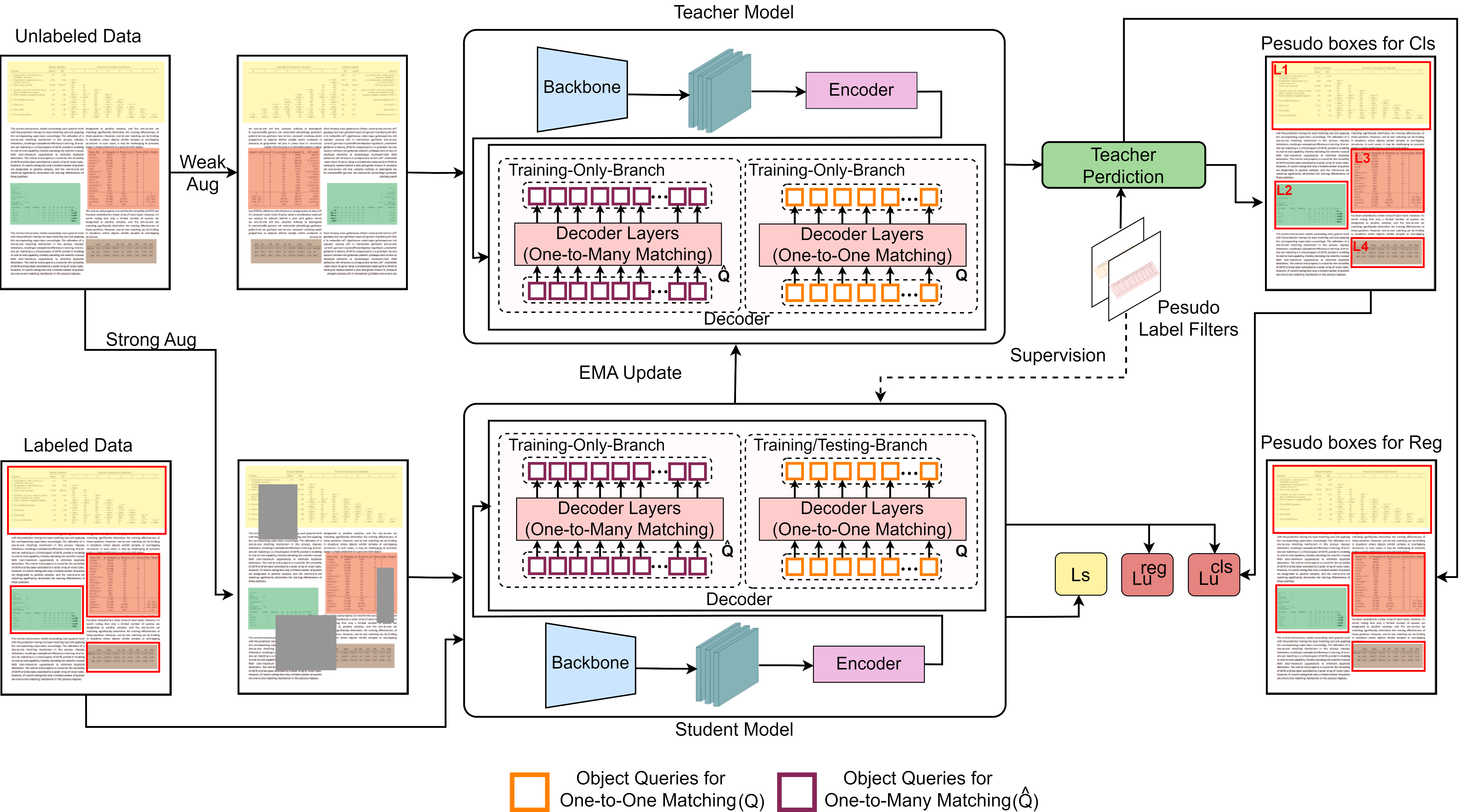}
    \captionsetup{font=small}
    \caption{\noindent\textbf{Our semi-supervised approach}: It focuses on employing one-to-many and one-to-one matching strategies in a semi-supervised setting and incorporates both labeled data and unlabeled data during training. The framework consists of two modules: student module and teacher module. The teacher module takes unlabeled images after employing weak augmentation prior and generates their pseudo-labels. The student module is designed to handle both labeled and unlabeled images, applying strong augmentation specifically to the unlabeled images. During the training process, the student module utilizes an EMA technique to consistently update the teacher module.}
     \label{fig: Figure: 3}
     \vspace{-1em} 
\end{figure}
In the one-to-many matching, the second set of queries, \(\hat{Q}\) go through the same \(L\) transformer decoder layers, resulting in \(L\) sets of predictions. To enable one-to-many matching \cite{fang2023feataug}, the ground-truth was replicated \(K\) times, leading to augment ground truths \(\hat{y_{g}}=\left\{ y_{g}^{1}, y_{g}^{2}, · · · , y_{g}^{K} \right\}\). Following this, bipartite matching is conducted between the predictions and the augmented ground truth pairs across each layer. The loss is then determined using the formula:
\vspace{-1em} 
\begin{equation}
\vspace{-0.8em} 
\mathcal{L}_{o2m} = \sum_{l=1}^{L} \mathcal{L}_{H}\left( \hat{y}_{p}, \hat{y}_{g}\right)
\end{equation}
Here, \(\hat{y}_{p}\) signifies the outcomes generated by the transformer decoder layer at position $l$. The Hungarian algorithm is employed for the proficient computation of the best assignment. The one-to-many matching strategy is exclusively applied during the model training phase. By integrating these matching strategies, the network improves overall detection performance. The one-to-many matching strategy provides high-quality predictions, while the one-to-one matching strategy removes duplicate predictions.

\subsection{Semi-Supervised Approach}
In this subsection, we introduce an improved semi-supervised approach for table detection in document images. For a detailed visual representation, please refer to Fig.~\ref{fig: Figure: 3}. Our approach contains two main modules: the student module and the teacher module. Both these modules contain transformer encoder-decoder network.
 During training, we have labeled image set \( D_{l} = \left\{ x^{l}_{i}, y^{l}_{i} \right\}^{N_{l}}_{i=1}\) and an unlabeled image set \( D_{u} = \left\{ x^{u}_{i} \right\}^{N_{u}}_{i=1} \),where \(N_{l}\) and \(N_{u}\) represents number of label and unlabel images, respectively.
The student module takes both labeled and unlabeled images as input, after applying strong augmentation on the unlabeled data and a combination of strong and weak augmentation on the labeled data. The teacher's module only fed unlabeled data with weak augmentation as input. The teacher network is responsible for generating pseudo-labels for the unlabeled data, which the student network then uses these pseudo-labels for training. 
Weak augmentation for the teacher module helps in generating better pseudo-labels for unlabeled data, while strong augmentation for the student module provides better learning process. At the start of the training, student module is trained on label data fed as input and update the teacher module using Exponential Moving Average (EMA) \cite{zhao2023does}. During the training process, teacher module generates pseudo-labels for unlabel data and fed to the student module. 

The decoder within these student-teacher modules utilizes two distinct matching strategies: one-to-many and one-to-one. This dual-strategy and collaborative framework between the student and teacher networks enhances the model's ability to learn from both labeled and unlabeled data, improving overall detection performance. By focusing on the quality and iterative refinement of pseudo-labels and leveraging the strengths of both matching strategies, our approach provides improved performance.

\noindent\textbf{Hybrid vs traditional matching:} Our primary contribution focuses on combining one-to-one and one-to-many matching strategies in a semi-supervised setting, advancing the core challenges of DETR-based Semi-Supervised table detection~\cite{shehzadi_semi-detr_table1}. Previous table detection approach with limited data employs a one-to-one matching strategy. When noisy pseudo labels generated in ~\cite{shehzadi_semi-detr_table1} employ a one-to-one matching strategy, it impacts the overall network performance. Our approach provides multiple positive proposals for each generated pseudo-label, filtering out noised pseudo-labels and selecting high-quality pseudo-labels. When fed to the student module, these high-quality pseudo-labels generated by the teacher module improve performance.

\noindent\textbf{Pseudo-labels for unlabeled data:} In the pseudo-labeling filtering framework, we filter out the pseudo-labels for unlabeled data generated by the teacher module. Object detectors typically produce a confidence score vector \(c_{k}\) for each bounding box \(b_{k}\). By applying a threshold to these scores, pseudo-labels are filtered out. 

In the early stages of semi-supervised table detection training, it is typical to encounter incorrect and unreliable pseudo-labels generated by the teacher module. The utilization of the one-to-one assignment approach has the potential for generating proposals that are both limited and of low quality. To improve the effectiveness of semi-supervised learning, we  employ the one-to-many assignment strategy as follows:
\begin{equation}
{\rho}_{o2m} = \left\{ \underset{\sigma_{i} \in \chi}{arg min} \sum_{i=1}^{N} \mathcal{L}_{H} \left( {y}_{i}^{T}, {y}_{\sigma_{i}\left( k \right)}^{S}  \right) \right\}_{i=1}^{\left| {y}^{T} \right|}
\end{equation}

Where \(\chi\) is used to represent the assignment of a subset of proposals to each pseudo label \({y}_{i}^{T}\). The term ${y}_{\sigma_{i}\left( k \right)}^{S}$ represents student generated prediction. In our approach, we use a one-to-many matching strategy in the teacher module to generate high-quality pseudo-labels ${y}_{i}^{T}$. The student module takes these pseudo-labels and trains the network for final predictions, employing one-to-many and one-to-one matching strategies. Achieving a one-to-one assignment involves implementing the Hungarian algorithm as follows:
\vspace{-1em} 
\begin{equation}
\vspace{-0.8em} 
{\rho}_{o2o} = \underset{\sigma\in \zeta_{N}}{arg min} \sum_{i=1}^{N} \mathcal{L}_{H} \left( {y}_{i}^{T}, {y}_{\sigma\left( i \right)}^{S}  \right)
\vspace{0.6em}
\end{equation}

Here, \(\zeta_{N}\) represents the collection of permutations involving $N$ elements. Additionally, \(\mathcal{L}_{H} \left( {y}_{i}^{T}, {y}_{\sigma\left( i \right)}^{S}  \right)\) stands for the cost incurred during the matching process between the pseudo labels ${y}^{T}$ and the prediction made by the student identified by the index $\sigma\left( i \right)$. 

\noindent\textbf{Losses:} The teacher model generates pseudo-label for unlabeled data, while the student model is trained on both labeled images with ground-truth annotations and unlabeled images with the pseudo-labels serving as the ground-truth. This setup allows for a combined loss function, which comprises a supervised loss \(\left( \mathcal{L}_{s} \right)\) from labeled images and an unsupervised loss \(\left( \mathcal{L}_{u} \right)\) from unlabeled images, weighted by a factor \(\omega\).

\vspace{-0.5em} 
\begin{equation}
\mathcal{L} = \mathcal{L}_{s} +\omega\mathcal{L}_{u}
\end{equation}

Initially, the model undergoes training with a one-to-many assignment strategy, where the loss is defined by the classification loss \(\left( \mathcal{L}_{o2m}^{cls} \right)\) and regression loss \(\left( \mathcal{L}_{o2m}^{reg} \right)\). 
\vspace{-0.5em} 
\begin{equation}
 \mathcal{L}_{o2m} = \mathcal{L}_{o2m}^{cls} +  \mathcal{L}_{o2m}^{reg}
\end{equation}

This approach leads to multiple positive proposals for each pseudo label, facilitating optimization and enhancing convergence speed. However, it also introduces duplicate predictions. To address this issue, the training switches to a one-to-one assignment strategy in the second stage. This transition retains the high-quality pseudo labels obtained from the initial training stage while gradually reducing duplicate predictions. Consequently, the loss functions in this stage remain consistent with the first stage, focusing on classification and regression losses.
\vspace{-0.5em} 
\begin{equation}
 \mathcal{L}_{o2o} = \mathcal{L}_{o2o}^{cls} +  \mathcal{L}_{o2o}^{reg}
\end{equation}
A significant contribution is our matching strategy, effectively addressing core challenges in DETR-based semi-supervised table detection~\cite{shehzadi_semi-detr_table1}. This leads to high-quality pseudo-labels, improving object detection accuracy. Furthermore, transitioning from one-to-one~\cite{shehzadi_semi-detr_table1} to one-to-many assignment enhances the performance of semi-supervised learning. The one-to-many assignment strategy for the whole training iterations increases training efficiency.  This refined approach assures a more robust and adaptable semi-supervised table detection system that improves overall performance without need NMS.
\section{Experimental Setup}
\label{sec:Experimental}

\subsection{Dataset and Evaluation Criteria}
\label{sec:Eval}
We use TableBank~\cite{tablebank8}, PubLayNet~\cite{PubLayNet3}, and ICDAR-19~\cite{icdar19} for evaluating our semi-supervised approach.
We utilize mean Average Precision (mAP), Average Precision at IoU threshold of 0.5 (AP\textsuperscript{50}), Average Precision at IoU threshold of 0.75 (AP\textsuperscript{75}), and Average Recall ( \(AR_{L}\)) to assess the effectiveness of our semi-supervised approach. These evaluation metrics collectively offer a comprehensive understanding of the model's performance in object detection.

\begin{figure}[ht]
    \centering
    \includegraphics[width=\linewidth, height=0.9\linewidth]{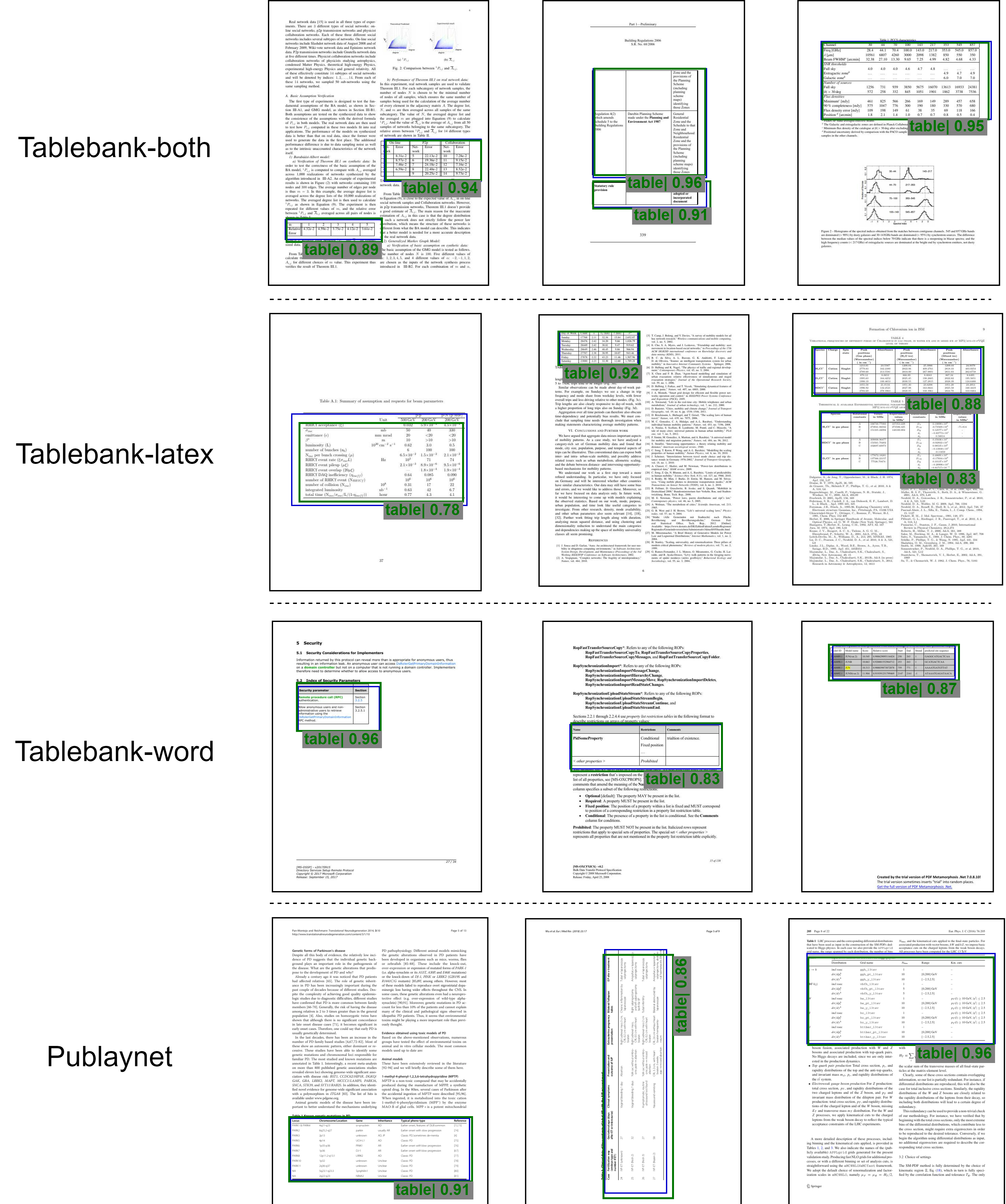}
    \captionsetup{font=small}
    \caption{Visualization of predictions of our approach on different datastets. It highlights that the inclusion of both one-to-one and one-to-many matching strategis enhance the model's accuracy. The ground truth is depicted by the blue boxes, while the green boxes display the results obtained through our approach.}
     \label{fig: Figure: 5}
     \vspace{-1em} 
\end{figure}

\subsection{Implementation Details}
\label{sec:implement}
We use the ResNet-50 \cite{resnet45} backbone, pre-trained on the ImageNet \cite{krizhevsky2012imagenet} dataset. We employ a one-to-one and one-to-many matching strategy for the decoder of the transformer network. We evaluate our approach on PubLayNet, ICDAR-19, and all three partitions of the TableBank dataset, utilizing 10$\%$, 30$\%$, and 50$\%$ of labeled data and rest as unlabeled data.  The pseudo-labeling threshold is set at 0.7. Training spans 150 epochs, with a learning rate reduction by a factor of 0.1 at the 140th epoch. Strong augmentations including horizontal flips, resizing, patch removal, cropping, grayscale, and Gaussian blur are applied, alongside weak augmentation through horizontal flipping. The optimal configuration for the number of queries to the decoder input for one-to-one matching is set at 30. Additionally, 400 queries for one-to-many matching are employed, and the ground truth K is replicated six times, all contributing to the optimal outcome. Unless otherwise specified, model evaluations employ mAP (AP50:95) metrics. The weights $\alpha_1$ and $\alpha_2$ for balancing the classification loss ($L_{cls}$) and regression loss ($L_{box}$) are set at 2 and 5, respectively. To improve training, input image dimensions are set to 600 pixels, with the standard size of 800 pixels retained for comparative purposes with other methodologies. Fig.~\ref{fig: Figure: 5} shows the performance of our semi-supervised method applied to the PubLayNet dataset and all splits of TableBank data, illustrating the efficiency of our approach across diverse document layouts.

\section{Results and Discussion}\label{sec:Results}

\subsection{TableBank}\label{subsubsec1}
\begin{figure}
    \centering
    \includegraphics[width=0.8\linewidth]{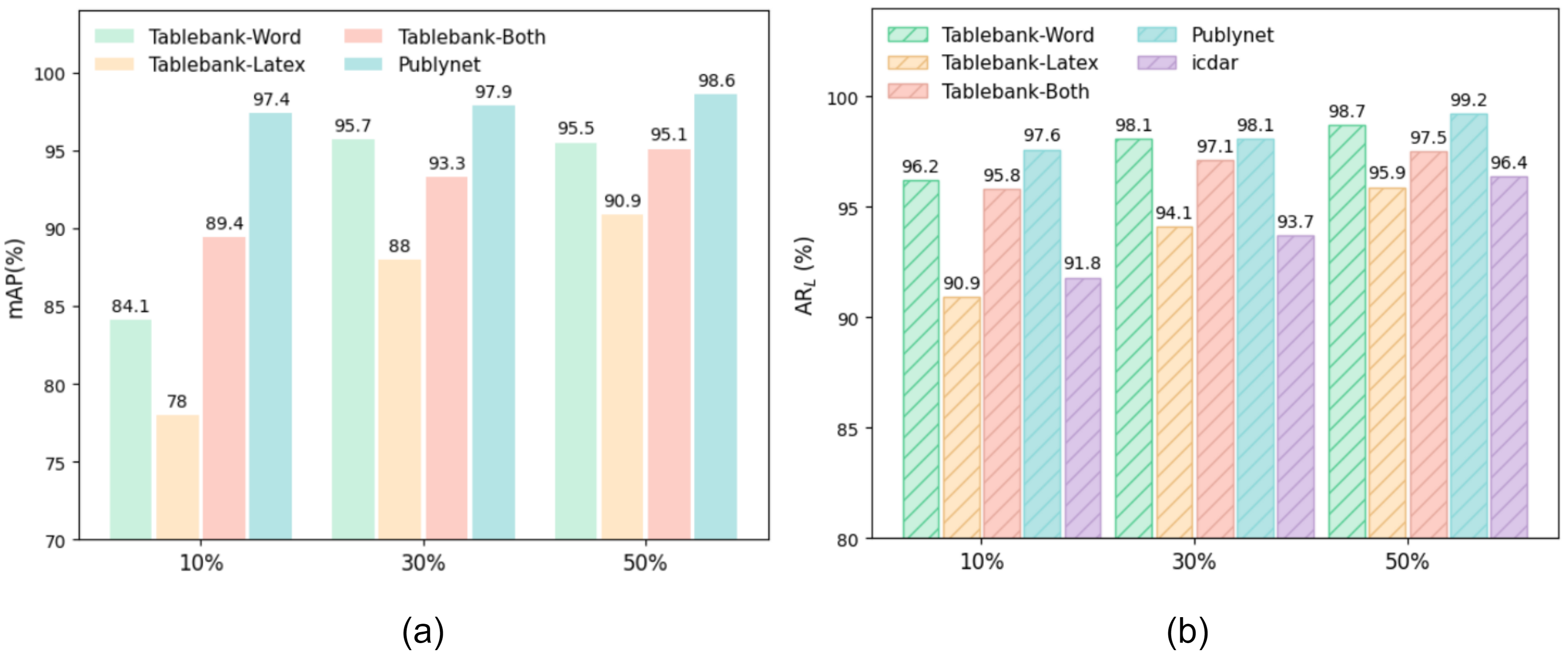}
    \captionsetup{font=small}
    \caption{Comparative performance analysis of our approach using the Tablebank, Publaynet, and ICDAR datasets. Experiments were conducted employing a ResNet-50 backbone with three distinct data splits: 10$\%$, 30$\%$, and 50$\%$. The results are presented as follows: (a) Mean Average Precision (mAP) within the IoU threshold range of 50$\%$ to 95$\%$, and (b) Average Recall for large objects (\(AR_{L}\)) within the IoU threshold range of 50$\%$ to 95$\%$.}
     \label{fig: Figure: 4}
     \vspace{-1em} 
\end{figure}

\begin{figure}
    \centering
    \includegraphics[width=0.8\linewidth, height=0.6\linewidth]{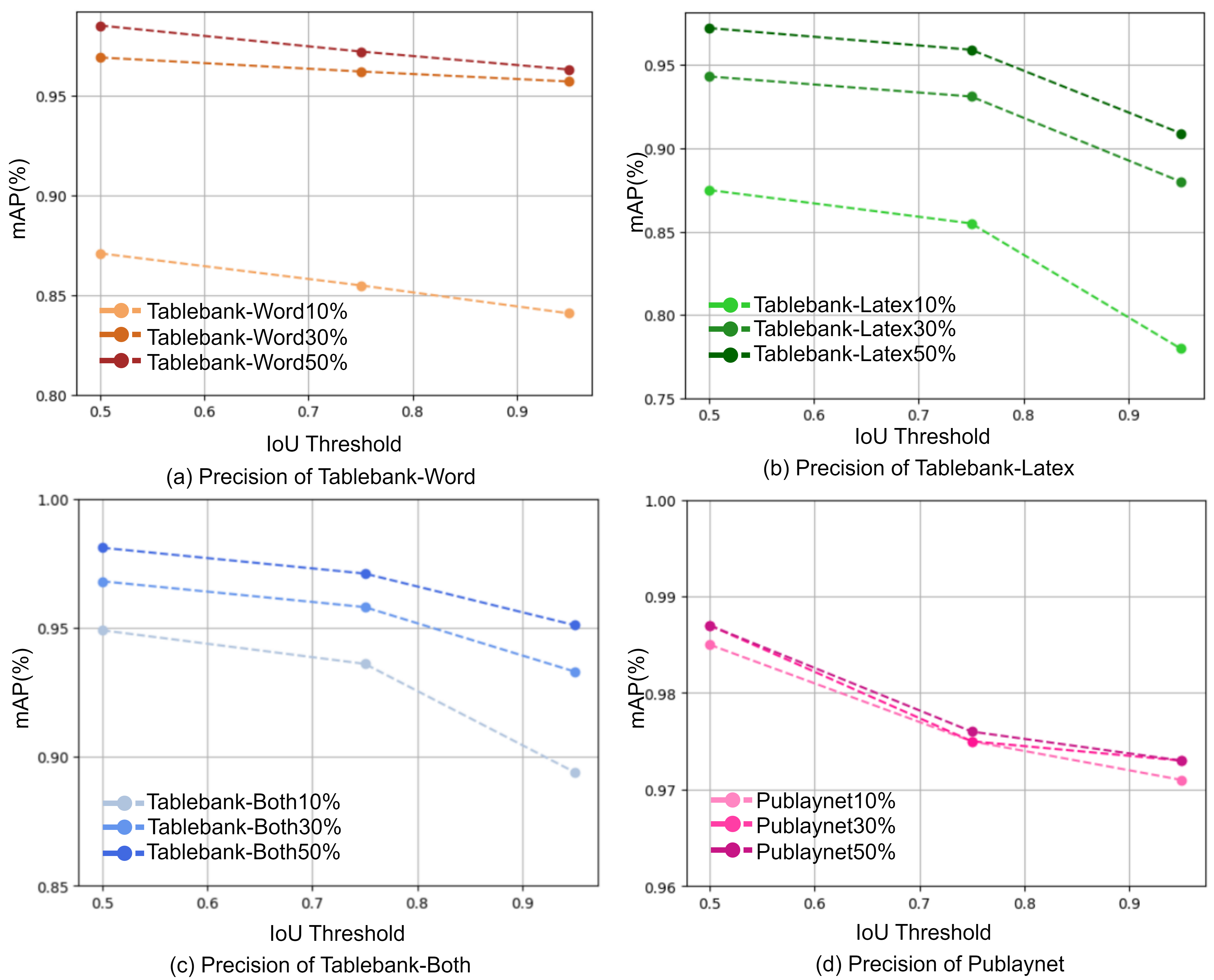}
    \captionsetup{font=small}
    \caption{Precision-Based Analysis of our Model's Performance.  This assessment covers four distinct datasets: (a) Tablebank-Word, (b) Tablebank-Latex, (c) Tablebank-Both and (d) Publaynet. Utilizing ResNet-50 as the foundational architecture, experiments were conducted across three different data splits: 10$\%$, 30$\%$, and 50$\%$.}
     \label{fig: Figure: 6}
     \vspace{-1.5em} 
\end{figure}

In this subsection, we present evaluate our semi-supervised table detection approach on all splits of the TableBank dataset \cite{tablebank8}, taking different proportions of labeled data. We also compare our transformer-based semi-supervised method, which utilizes one-to-many and one-to-one assignment strategies, against previous supervised and semi-supervised approaches.
Table~\ref{tab:tableno1} shows the results obtained from our approach, which employs the deformable transformer~\cite{Deformable54} across the TableBank-word, TableBank-latex, and TableBank-both data splits. These experiments uses 10$\%$, 30$\%$, and 50$\%$ labeled data, with the remaining as unlabeled data. Notably, with 50\% labeled data, the TableBank-word, TableBank-latex, and TableBank-both split achieve mAP scores of 95.5\%, 90.9\%, and 95.1\%, respectively, showcasing the efficiency of our semi-supervised method. Moreover, our approach shows AP\textsuperscript{50} of 98.5, 97.2, and 98.1 at 50\% label data on TableBank-word, TableBank-latex, and TableBank-both splits, respectively. Figure~\ref{fig: Figure: 4}~(a) illustrates the mAP achieved by our semi-supervised method on the PubLayNet and TableBank datasets at various percentages of labeled data. Figure~\ref{fig: Figure: 4}~(b) depicts the \(AR_L\) for the PubLayNet, TableBank, and ICDAR-19 datasets, also across differing levels of labeled data. Figure~\ref{fig: Figure: 6} demonstrates the mAP at different Intersection over Union (IoU) thresholds for the TableBank-Word, TableBank-Latex, TableBank-Both, and PubLayNet datasets. Each dataset is analyzed at three percentage of labeled data: 10\%, 30\%, and 50\%.

\begin{table*}[h!]
\begin{center}
\renewcommand{\arraystretch}{1} 
\centering
\begin{tabular*}{\textwidth} 
{@{\extracolsep{\fill}}llllll@{\extracolsep{\fill}}}
\toprule
\textbf{Dataset} &
\textbf{Label-splits} &
\textbf{mAP} & 
\textbf{AP\textsuperscript{50}} &
\textbf{AP\textsuperscript{75}}  & 
\textbf{AR\textsubscript{L}}  \\
\toprule
\multirow{3}{*}{TableBank-word }  & 10$\%$  & 84.1 & 87.1 & 85.5 & 96.2  \\
& 30$\%$ & 95.7 & 96.9 & 96.2 & 98.1  \\
& 50$\%$ & 95.5 & 98.5 & 97.2 & 98.7 \\
\midrule
 \multirow{3}{*}{TableBank-latex } & 10$\%$ & 78.0 & 87.5 & 85.5 & 90.9  \\
 & 30$\%$ & 88.0  & 94.3 & 93.1 & 94.1  \\
 & 50$\%$ & 90.9 & 97.2 & 95.9 & 95.9 \\
 \midrule
\multirow{3}{*}{TableBank-both }  & 10$\%$  & 89.4 & 94.9 & 93.6 & 95.8    \\
 & 30$\%$ & 93.3 & 96.8 & 95.8 & 97.1 \\
 & 50$\%$ & 95.1 & 98.1 & 97.1 & 97.5  \\
\bottomrule
\end{tabular*}
\captionsetup{font=small}
\caption{Evaluation of our semi-supervised approach across TableBank-word, TableBank-latex, and
TableBank-both data divisions with varying proportions of labeled data. Here, mAP denotes the mean Average Precision within the IoU range of 0.50 to 0.95, \(AP^{50}\) signifies the Average Precision at an
IoU threshold of 0.5, and \(AP^{75}\) represents the Average Precision at an IoU threshold of 0.75. Additionally, \(AR_{L}\)  refers to the Average Recall for larger objects.}\label{tab:tableno1}
\end{center}
\vspace{-2em} 
\end{table*}

\begin{table}[h!]
\centering
\begin{tabular*}{\textwidth} 
{@{\extracolsep{\fill}}llllll@{\extracolsep{\fill}}}
\toprule%
\textbf{Method} &
\textbf{Technique} &
\textbf{Detector} & 
\textbf{10$\%$} &
\textbf{30$\%$} &
\textbf{50$\%$} \\
\midrule
Ren et al. \cite{ren2015faster}. & Sup & Faster R-CNN & 80.1 & 80.6 & 83.3 \\
\midrule
Zhu et al.\cite{Deformable54}. & Sup & Def. DETR & 80.8 & 82.6 & 86.9 \\
\midrule
Sohn et al. \cite{simplesemi76} & Semi-sup & Faster R-CNN & 82.4 & 83.8 & 87.1 \\
\midrule
Liu et al. \cite{liu2021unbiased} & Semi-sup & Faster R-CNN & 83.9 & 86.4 & 88.5 \\
\midrule
Tang et al. \cite{tang2021humble} & Semi-sup & Faster R-CNN & 83.4 & 86.2 & 87.9 \\
\midrule
Xu et al. \cite{softTeacher56} & Semi-sup & Faster R-CNN & 83.6 & 86.8 & 89.6 \\
\midrule
Shehzadi et al.\cite{shehzadi_semi-detr_table1} & Semi-sup & Def. DETR & 84.2 & 86.8 & 91.8 \\
\midrule
Our & Semi-sup & Def. DETR & 89.4 & 93.3 & 95.1 \\
\botrule
\end{tabular*}
\captionsetup{font=small}
\caption{Comparison of the performance of previous supervised and semi-supervised methodologies.  The ResNet-50 backbone was used to train the supervised models, Deformable-DETR(Def. DETR) and Faster R-CNN, using 10$\%$, 30$\%$, and 50$\%$ of the TableBank-both dataset. The same splits and datasets are utilized for semi-supervised assessment. It's important to note that all results are given in terms of mAP (0.5:0.95).}\label{tab:tableno2}
\vspace{-1.5em} 
\end{table}

In Table~\ref{tab:tableno2}, we present a comparison between our semi-supervised approach and earlier supervised as well as semi-supervised methods, showing results with mAP scores trained on 10\%, 30\%, and 50\% labeled data. Our method stands out by implementing both one-to-many and one-to-one assignment strategies, which generate superior pseudo-labels for unlabeled data. Notably, on just 10\% labeled data, our semi-supervised approach outperforms the previous supervised baseline network by 8.6 mAP points and the prior semi-supervised method by 5.2 mAP points.

\subsection{PubLayNet}\label{subsubsec2}
In this subsection, we delve into the experimental results for table detection in the PubLayNet dataset~\cite{PubLayNet3}, exploring the impact of different amounts of labeled data. We also compare our transformer-based semi-supervised method with previous supervised and semi-supervised methods, offering a detailed evaluation of their performance.
\begin{table*}[h]
\begin{center}
\renewcommand{\arraystretch}{1} 
\begin{tabular*}{\textwidth}
{@{\extracolsep{\fill}}lcllll@{\extracolsep{\fill}}}
\toprule
\textbf{Dataset} &
\textbf{Label-percent} &
\textbf{mAP} & 
\textbf{AP\textsuperscript{50}} &
\textbf{AP\textsuperscript{75}} &
\textbf{AR\textsubscript{L}}\\
\toprule
 \multirow{3}{*}{publaynet} & 10$\%$ & 97.4 & 98.5 & 98.3 & 97.6\\
 & 30$\%$ & 97.9  & 98.7 & 98.5 & 98.1  \\
 & 50$\%$ & 98.6 & 98.9 & 99.3 & 99.2  \\
 \midrule
\end{tabular*}
\captionsetup{font=small}
\caption{Evaluation results for the table class in the PubLayNet dataset using three different data splits: 10$\%$, 30$\%$, and 50$\%$.}\label{tab:tableno3}
\end{center}
\vspace{-1em} 
\end{table*} 

The results of our semi-supervised approach for table detection in PubLayNet are presented in Table~\ref{tab:tableno3}, with performance measured across various quantities of labeled data. Notably, the mAP values achieved are 97.4\%, 97.9\%, and 98.6\% for our semi-supervised approach trained with 10\%, 30\%, and 50\% labeled data, respectively.

\begin{table}[h!]
    \centering
    \begin{tabular*}{\textwidth}
{@{\extracolsep{\fill}}lcccccl@{\extracolsep{\fill}}}

\toprule
\textbf{Method} & \multicolumn{3}{c}{\textbf{mAP}} & \multicolumn{3}{c}{\textbf{AR\textsubscript{L}}} \\
\cmidrule(lr){2-4}\cmidrule(lr){5-7}
           & \textbf{10$\%$} & \textbf{30$\%$} & \textbf{50$\%$} & \textbf{10$\%$} & \textbf{30$\%$} & \textbf{50$\%$} \\
\midrule
\text{Shehzadi et al.\cite{shehzadi_semi-detr_table1}}  & 88.4 & 90.3 & 92.8 & 91.0 & 93.2 & 96.0 \\
\text{Ours} & 97.4 & 97.9 & 98.6 & 97.6 & 98.1 & 99.2 \\
\bottomrule
\end{tabular*}
\captionsetup{font=small}
    \caption{Comparison of our approach with previous table detection semi-supervised approach on PubLayNet data using the splits of 10$\%$, 30$\%$, and 50$\%$ of labeled data.} \label{tab:tableno4}
\label{tab:semi-comp}
\end{table}

In Table~\ref{tab:semi-comp}, we compare our semi-supervised approach with previous semi supervised table detection appraoch. For mAP, 
with 10\% labeled data, the improvement is 8.9 percentage points (from 88.4\% to 97.4\%), at 30\% it's 7.6 points (from 90.3\% to 97.9\%), and at 50\% the improvement is 5.8 points (from 92.8\% to 98.6\%). For \(AR_{L}\), the gains are even more significant: with 10\% labeled data, the improvement is 6.6 points (from 91.0\% to 97.6\%), at 30\% it's 5.9 points (from 93.2\% to 98.1\%), and at 50\% it's a substantial 3.2 point increase (from 96.0\% to 99.2\%).

 \begin{table}[h!]
\begin{tabular*}{\textwidth}{@{\extracolsep\fill}lcccccc}
\toprule%
\textbf{Method} &
\textbf{Technique} &
\textbf{Detector} & 
\textbf{10$\%$} &
\textbf{30$\%$} &
\textbf{50$\%$} \\
\toprule
Ren et al. \cite{ren2015faster} & Sup & Faster R-CNN & 93.6 & 95.6 & 95.9 \\
\midrule
Zhu et al. \cite{Deformable54}  & Sup & Def. DETR & 93.9 & 96.2 & 97.1 \\
\midrule
Sohn et al. \cite{simplesemi76} & Semi-sup & Faster R-CNN & 95.8 & 96.9 & 97.8 \\
\midrule
Liu et al.\cite{liu2021unbiased} & Semi-sup & Faster R-CNN & 96.1 & 97.4 & 98.1 \\
\midrule
Tang et al.\cite{tang2021humble} & Semi-sup & Faster R-CNN & 96.7 & 97.9 & 98.0 \\
\midrule
Xu et al. \cite{softTeacher56}  & Semi-sup & Faster R-CNN & 96.5 & 98.1 & 98.5 \\
\midrule
Our & Semi-sup & Def. DETR & 98.5 & 98.7 & 98.9 \\
\botrule
\end{tabular*}
\captionsetup{font=small}
\caption{Performance comparison with earlier supervised and semi-supervised approaches. Here, Deformable-DETR is represented as Def. DETR. All the supervised and unsupervised networks are trained on 10$\%$, 30$\%$, and 50$\%$ of PubLayNet table class data. The results are presented in terms of \(AP{50}\) at an IoU threshold of 0.5.}\label{tab:tableno5}
\vspace{-1.5em} 
\end{table}

Table~\ref{tab:tableno5} presents a comprehensive evaluation of both supervised and semi-supervised networks with specific emphasis on the table class from the PubLayNet dataset \cite{PubLayNet3}. We additionally compare a supervised deformable-DETR model, which was trained on varying proportions of labeled data (10$\%$, 30$\%$, and 50$\%$) from the PubLayNet table class, with our semi-supervised approach employing the deformable transformer \cite{Deformable54}. Notably, our methodology produces comparable outcomes, eliminating the need for post-processing steps like NMS.

\subsection{ICDAR-19}\label{subsubsec3}
We also evaluate our semi-supervised approach on ICDAR-19 \cite{icdar19} Modern Track A dataset. Table~\ref{tab:tableno6} shows the results of our approach on the ICDAR-19 dataset, utilizing 10$\%$, 30$\%$, and 50$\%$ of annotated data. Table~\ref{tab:tableno7} provides a comprehensive comparison of our approach with previous supervised and semi-supervised networks at IOU thresholds of 0.8 and 0.9.
\begin{table*}[h]
\begin{center}
\renewcommand{\arraystretch}{1} 
\begin{tabular*}{\textwidth}
{@{\extracolsep{\fill}}llllll@{\extracolsep{\fill}}}
\toprule
\textbf{Dataset} &
\textbf{Label-percent} &
\textbf{mAP} & 
\textbf{AP\textsuperscript{50}} &
\textbf{AP\textsuperscript{75}}  & 
\textbf{AR\textsubscript{L}}  \\
\toprule
\multirow{3}{*}{icdar }  & 10$\%$  & 60.2 & 64.7 & 61.4 & 91.8  \\
& 30$\%$ & 64.8 & 69.0 & 65.6 & 93.7  \\
& 50$\%$ & 65.4 & 68.6 & 67.1 & 96.4 \\
\midrule
\end{tabular*}
\captionsetup{font=small}
\caption{Evaluation results for the ICDAR-19 dataset using three different data splits: 10$\%$, 30$\%$, and 50$\%$.}\label{tab:tableno6}
\end{center}
\vspace{-2em}
\end{table*}

\begin{figure}
    \centering
    \includegraphics[width=0.8\linewidth]{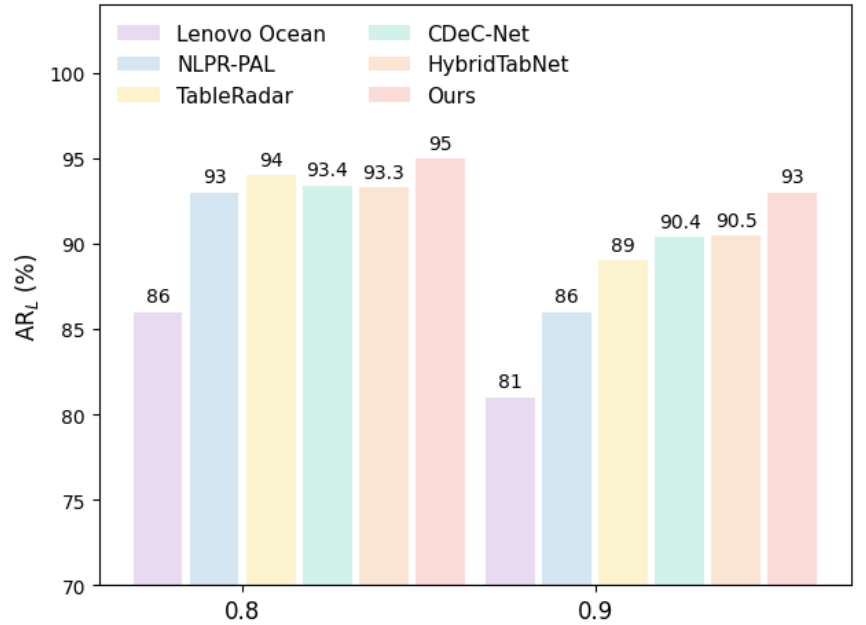}
    \captionsetup{font=small}
    \caption{Comparing our approach with other supervised approaches based on recall on the ICDAR 19 Track A dataset with IOU thresholds of 0.8 and 0.9.}
     \label{fig: Figure: 7}
     \vspace{-1.5em} 
\end{figure}
 These approaches are evaluated based on Recall, Precision, and F1-Score. Our approach shows a high recall for table detection, indicating its strength in identifying nearly all actual tables at both IOU thresholds of 0.8 and 0.9.
\begin{table}[h!]
    \centering
    \begin{tabular*}{\textwidth}
{@{\extracolsep{\fill}}llcccccl@{\extracolsep{\fill}}}

\toprule
\textbf{Method} & \textbf{Approch} & \multicolumn{3}{c}{$\text{IOU}=\text{0.8}$} & \multicolumn{3}{c}{$\text{IOU}=\text{0.9}$} \\
\cmidrule(lr){3-5}\cmidrule(lr){6-8}
          & & Recall & Precision & F1-Score & Recall & Precision & F1-Score \\
\midrule
\text{TableRadar\cite{icdar19}} &\text{Sup}   & 94.0 & 95.0 & 94.5 & 89.0 & 90.0 & 89.5 \\
\text{NLPR-PAL\cite{icdar19}} &\text{Sup} & 93.0 & 93.0 & 93.0 & 86.0 & 86.0 & 86.0 \\
\text{Lenovo Ocean\cite{icdar19}} &\text{Sup} & 86.0 & 88.0 & 87.0 & 81.0 & 82.0 & 81.5\\
\text{CDeC-Net\cite{Agarwal52}} &\text{Sup} & 93.4 & 95.3 & 94.4 & 90.4 & 92.2 & 91.3\\
\text{HybridTabNet\cite{Hyb65}} &\text{Sup} & 93.3 & 92.0 & 92.8 & 90.5 & 89.5 & 90.2 \\
\text{Shehzadi et al.\cite{shehzadi_semi-detr_table1}} &\text{Semi-sup} & 71.1 & 82.3 & 76.3 & 66.3 & 76.8 & 71.2 \\
\text{Ours} &\text{Semi-sup} & 95.0 & 61.4 & 74.5 & 93.0 & 60.7 & 73.4 \\
\bottomrule
\end{tabular*}
\captionsetup{font=small}
    \caption{ Performance comparison between the proposed semi-supervised approach (with 50\% labels) and previous state-of-the-art results on the ICDAR 19 Track A dataset with IOU thresholds of 0.8 and 0.9. } \label{tab:tableno7}
 \vspace{-1.5em}
\end{table}
However, this comes with lower precision, suggesting it also picks up some false positives. Despite this, the approach effectively ensures that few actual tables are missed. Moreover, Fig.~\ref{fig: Figure: 7} shows the recall results of a comparison of our approach with the previous approach on the ICDAR-19 dataset. Here, we can observe that our approach shows the highest recall results.
\vspace{-1em} 
\begin{table*}[htp!]
\begin{center}
\renewcommand{\arraystretch}{1.1} 
\begin{tabular*}{\textwidth}
{@{\extracolsep{\fill}}ccccccc@{\extracolsep{\fill}}}
\toprule
\textbf{Method} & 
\textbf{Technique} & 
\textbf{ Labels} &
\textbf{TableBank} &
\textbf{PubLayNet } \\
\toprule
CDeC-Net \cite{Agarwal52} & Sup & 100\%  &  96.5 & 97.8 \\
\midrule
EDD \cite{zhong2020image} & Sup & 100\%  &  86.0  & 89.9 \\
\midrule
TableFormer \cite{nassar2022tableformer} & Sup & 100\%  &  89.6 & 96.7 \\
\midrule
CasTabDetectoRS \cite{CasTab45} & Sup & 100\% & 95.3 & - \\
\midrule
VSR \cite{vsr45} & Sup & 100\% & - & 95.9\\
\midrule
Shehzadi et al.\cite{shehzadi_semi-detr_table1}  & Semi-sup & 30\%  & 86.8 & 90.3 \\
\midrule
Our & Semi-sup & 30\%  & 93.3 & 97.9\\
\bottomrule
\end{tabular*}
\captionsetup{font=small}
\caption{Performance comparison with previous supervised and semi-supervised table detection methods on TableBank-both and PubLaynet data. The results are the mean average precision (mAP) across the range from 0.5 to 0.95.} \label{tab:tableno8}
\end{center}
\end{table*}
\vspace{-2em}

In addition, we conducted a comparative analysis of our semi-supervised method using 30$\%$ labeled data against earlier techniques in table detection and document analysis on TableBank and PubLayNet datasets (refer to Table~\ref{tab:tableno8}). Notably, we observe that even with only 30$\%$ labeled data, our approach yielded results comparable to those obtained by previous supervised methods. 
\section{Ablation Study}\label{sec:Ablation}
In this subsection, we evaluate the main components of our approach. All ablation studies utilize a ResNet-50 backbone along with a dataset subset consisting of 10$\%$ labeled images of PubLayNet data.

\noindent\textbf{Impact of the quantity of learnable queries}: We examine the effect of the quantity of queries fed to the decoder. Optimal performance is achieved when employing 30 queries in a one-to-one and 400 queries in a one-to-many assignment strategy. Deviating from these values leads to a decline in performance, as observed in ~Table~\ref{tab:tableno12}. We selected the 30 queries in one-to-one assignment as in the baseline; it also give the best results in our case. We then vary the queries in one-to-many assignment strategy and get the best results at 400. Deviating from this results in a performance decrease due to overfitting and overlapping predictions. 
\begin{table*}[h!]
\begin{center}
\renewcommand{\arraystretch}{1} 
\centering
\small 
\begin{tabular*}{\linewidth} 
{@{\extracolsep{\fill}}cccccc@{\extracolsep{\fill}}}

\toprule
\textbf{Method} &
\textbf{$\#$ query} &
\textbf{mAP} & 
\textbf{AP\textsuperscript{50}} &
\textbf{AP\textsuperscript{75}}  & 
\textbf{AR\textsubscript{L}}  \\
\toprule
\multirow{1}{*}{Shehzadi et al. \cite{shehzadi_semi-detr_table1} }  & o2o=30,o2m=0   & 88.4 & 98.5 & 97.3 & 91.0  \\
\midrule
 \multirow{3}{*}{Ours}  & o2o=30,o2m=200 & 89.8  & 95.2 & 94.2 & 93.4  \\
 &\textbf{o2o=30,o2m=400} & \textbf{97.4} & \textbf{98.5} & \textbf{98.3} & \textbf{97.6}  \\
 
 & o2o=30,o2m=600 & 90.9 & 97.2 & 95.9 & 95.9 \\
 
\bottomrule
\end{tabular*}
\captionsetup{font=small}
\caption{Comparison of our approach with semi-supervised deformable DETR (Def. DETR) based on varying numbers of queries as a hyperparameter.}\label{tab:tableno12}
\end{center}
 \vspace{-1.5em}
\end{table*}

\noindent\textbf{Pseudo-labeling confidence threshold}: The filtering threshold is a critical parameter influencing the quality and quantity of pseudo-labels filtered out during semi-supervised learning. Increasing the threshold reduces the number of pseudo-labels fed to the student module, but those are of higher quality. Conversely, lowering the threshold leads to more pseudo-labels filtered out, which also increases noisy pseudo-labels. Table~\ref{tab:table8} presents a comprehensive summary of the effects observed across several threshold values, from 0.5 to 0.8. According to the results, the most suitable threshold value is 0.7.

\begin{table*}[h]
\vspace{-1em} 
\begin{minipage}{0.4\textwidth}
       \small
       \begin{threeparttable}
        \begin{tabular*}{0.9\textwidth}
{@{\extracolsep{\fill}}llll@{\extracolsep{\fill}}}

\toprule
\textbf{T} &
\textbf{mAP} & 
\textbf{AP\textsuperscript{50}} &
\textbf{AP\textsuperscript{75}}  \\
\toprule
0.5  & 96.9 & 98.3 & 97.2  \\
\hline
0.6  & 96.2 & 97.6 & 96.0   \\
\hline
\textbf{0.7}  & \textbf{97.4} & \textbf{98.5} & \textbf{98.3}  \\
\hline
0.8  & 96.8 & 98.3 & 97.2  \\
\midrule
\end{tabular*}
\captionsetup{font=small}
        \caption{Analysis of pseudo-labeling confidence thresholds and their performance impact, highlighting top-performing results in bold.}\label{tab:table8}
        \end{threeparttable}
    \end{minipage}%
\begin{minipage}{0.65\textwidth}
        \small
        \begin{threeparttable}
            \begin{tabular*}{0.9\textwidth}
{@{\extracolsep{\fill}}lllll@{\extracolsep{\fill}}}

\toprule
\textbf{Method} &
\textbf{mAP} & 
\textbf{AP\textsuperscript{50}} &
\textbf{AP\textsuperscript{75}}  & 
\textbf{AR\textsubscript{L}}  \\
\toprule
\multirow{1}{*}{H-DETR \cite{jia2023detrs} }  & 95.2 & 97.7 & 96.8 & 98.5  \\
\midrule
 \multirow{1}{*}{Co-DETR\cite{zong2023detrs} }  & 96.3 & 98.2 & 97.6 & 98.9  \\
 \midrule
 \multirow{1}{*}{Ours} & \textbf{97.4} & \textbf{98.5} & \textbf{98.3} & \textbf{97.6}  \\
\midrule
\end{tabular*}
\captionsetup{font=small}
        \caption{Comparison of our approach with other models using PubLayNet, employing a one-to-many assignment strategy to enhance object detection performance without the use of NMS.} \label{tab:table9}
        \end{threeparttable}
    \end{minipage}
    \vspace{-1.5em} 
\end{table*}

\begin{figure}
    \centering
    \includegraphics[width=0.8\linewidth, height=0.6\linewidth]{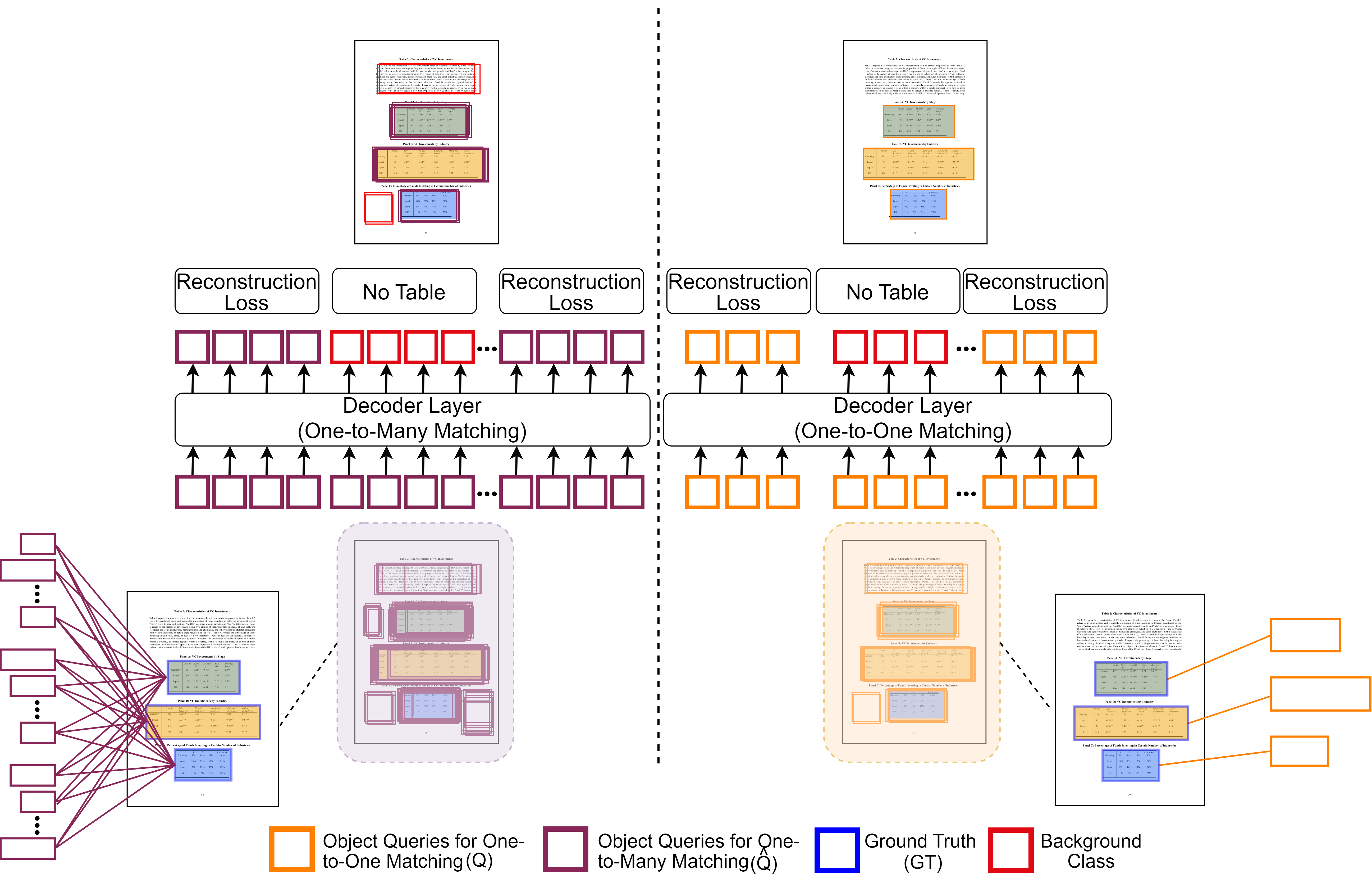}
    \captionsetup{font=small}
    \caption{Performance analysis conducted by inputting 30 one-to-one queries and 400 one-to-many queries into the decoder layer. Purple rectangles represent one-to-many queries, orange rectangles denote one-to-one queries, blue rectangles indicate ground-truth and red rectangles signify the background class.}
     \label{fig: Figure: 8}
     \vspace{-1.5em} 
\end{figure}

\noindent\textbf{Impact of one-to-one and one-to-many matching strategies} 
In one-to-one matching, each ground truth matches the most similar prediction. On the other hand, one-to-many matching, as shown in Fig.~\ref{fig: Figure: 8}, allows multiple predictions to match with a single ground truth, enhancing training but potentially causing overlapping predictions during testing. To balance these factors, our strategy involves using one-to-one and one-to-many matching during training and exclusively one-to-one matching during testing to ensure accurate predictions without the NMS step. Table~\ref{tab:table9} compares our semi-supervised one-to-many matching strategy with previous supervised one-to-many matching strategies. In Table~\ref{tab:tableno11}, We also observe the effect of one-to-one and one-to-many matching strategies. One-to-one matching takes less training time as it doesn't need an NMS step but shows low performance. One-to-many matching needs longer training time as an NMS step is needed to remove duplicate predictions. We first employ one-to-many matching to improve performance by selecting high-quality pseudo-labels and then shift to one-to-one matching to remove duplicate predictions. Our approach does not need NMS and improves performance.

\begin{table}[h!]
    \centering
    \begin{tabular*}{\textwidth}
{@{\extracolsep{\fill}}cccccccccc@{\extracolsep{\fill}}}

\toprule
\textbf{o2m} &
\textbf{o2o} &
\textbf{NMS} & 
\textbf{Training Time(h) }&
\textbf{FPS }&
\textbf{mAP} & 
\textbf{AP\textsuperscript{50}} &
\textbf{AP\textsuperscript{75}} &
\textbf{AR\textsubscript{L}} \\
\toprule
\color{black}\xmark   &  \checkmark & \color{black}\xmark  &8.74 & 4.36& 96.0 & 97.8 & 96.8  &98.9\\
\hline
\checkmark  &  \color{black}\xmark   & \checkmark  & 11.33 & 4.36& 95.5 & 96.7 & 96.5 &97.5\\
\hline
\multirow{1}{*} {\checkmark} & \multirow{1}{*} {\checkmark}  & \multirow{1}{*}{\color{black}\xmark}   & \textbf{10.99}  & \textbf{4.36} & \textbf{96.1} & \textbf{97.9} & \textbf{96.9} & \textbf{99.0} \\

\hline
\end{tabular*}
\captionsetup{font=small}
    \caption{Comparative analysis of one-to-many(o2m) and one-to-one(o2o) matching strategies. performance metrics, including NMS-based matching, training time in hours, frames per second (FPS), These are the results without using augmented GT in one-to-many matching strategy. } \label{tab:tableno11}
 \vspace{-1.5em}
\end{table}

\section{Conclusion}
\label{sec:Conclusion}
In conclusion, our study addresses the challenges inherent in DETR-based object detectors within semi-supervised object detection. We recognized the limitations of the conventional one-to-one assignment approach, particularly its inefficiency in handling imprecise pseudo-labels. To overcome these issues, we introduced a semi-supervised, transformer-based approach for semi-supervised object detection that operates end-to-end. Our method integrates one-to-one and one-to-many matching strategies, improving overall performance. We generate the pseudo-labels by employing two distinctive modules, student and teacher. Through the iterative update process facilitated by an Exponential Moving-Average (EMA) function, these modules enhanced one another, resulting in accurate classification and bounding box predictions. The experimental findings shows the effectiveness of our approach, surpassing the performance of supervised and semi-supervised approaches across various labeling ratios on TableBank and the PubLayNet training data. In future, we aim to extend the semi-supervised learning framework based on transformers to tackle the task of table structure recognition. This research represents significant progress in advancing more robust and efficient methodologies for semi-supervised object detection within document analysis tasks.


\bibliography{mybib}

\end{document}